\documentclass[sigconf]{acmart}
\usepackage{multirow, dsfont}

\AtBeginDocument{%
  }

\copyrightyear{2025}
\acmYear{2025}
\setcopyright{acmlicensed}\acmConference[KDD '25]{Proceedings of the 31st ACM SIGKDD Conference on Knowledge Discovery and Data Mining V.1}{August 3--7, 2025}{Toronto, ON, Canada}
\acmBooktitle{Proceedings of the 31st ACM SIGKDD Conference on Knowledge Discovery and Data Mining V.1 (KDD '25), August 3--7, 2025, Toronto, ON, Canada}
\acmDOI{10.1145/3690624.3709408}
\acmISBN{979-8-4007-1245-6/25/08}

\begin{document}

\title{Multi-Branch Collaborative Learning Network for Video Quality Assessment in Industrial Video Search}

\renewcommand{\shorttitle}{MBCN}
\author{Hengzhu Tang}
\affiliation{
  \institution{Baidu Inc.}
  \city{Beijing}
  \country{China}
 }
\email{hengzhutang@gmail.com}

\author{Zefeng Zhang}
\affiliation{
  \institution{Baidu Inc.}
  \city{Beijing}
  \country{China}
 }
\email{zhangzefeng02@baidu.com}

\author{Zhiping Li}
\affiliation{
  \institution{Baidu Inc.}
  \city{Beijing}
  \country{China}
 }
\email{lizhiping07@baidu.com}

\author{Zhenyu Zhang}
\affiliation{
  \institution{Baidu Inc.}
  \city{Beijing}
  \country{China}
 }
\email{zhangzhenyu07@baidu.com}

\author{Xing Wu}
\affiliation{
  \institution{Baidu Inc.}
  \city{Beijing}
  \country{China}
 }
\email{wuxing04@baidu.com}

\author{Li Gao}
\affiliation{
  \institution{Baidu Inc.}
  \city{Beijing}
  \country{China}
 }
\email{gaoli.sinh@gmail.com}
\authornote{Corresponding author.}

\author{Suqi Cheng}
\affiliation{
  \institution{Baidu Inc.}
  \city{Beijing}
  \country{China}
 }
\email{chengsuqi@gmail.com}

\author{Dawei Yin}
\affiliation{
  \institution{Baidu Inc.}
  \city{Beijing}
  \country{China}
 }
\email{yindawei@acm.org}

\renewcommand{\shortauthors}{Hengzhu Tang et al.}

\begin{abstract}

Video Quality Assessment (VQA) is a crucial component of broadscale video retrieval systems. Its goal is to accurately identify various quality issues in videos, thereby encouraging the video retrieval system to prioritize high-quality videos.
In large-scale industrial video retrieval systems, we formulate the characteristics of low-quality videos into four categories: visual-related low-level quality problems such as mosaics and black boxes, textual-related low-level quality problems caused by video title and Optical Character Recognition (OCR) content, as well as semantic-level frame incoherence and frame-text mismatch caused by emerging AI-generated videos. 
These kinds of low-quality videos, which are widely present in industrial environments, have been overlooked in academic research before, and accurately identifying them is very challenging.
In this paper, we introduce a \textbf{M}ulti-\textbf{B}ranch \textbf{C}ollaborative learning \textbf{N}etwork (\textbf{MBCN}) to tackle the above issues.
We carefully design four assessment branches for MBCN to adapt to the above four kinds of issues for industrial video retrieval systems. After obtaining independent scores for each branch, we perform a weighted aggregation of the various branches to dynamically address video quality issues in different scenarios with a squeeze-and-excitation mechanism.
Finally, we integrate point-wise and pair-wise optimization objectives to ensure the predicted scores are stable and fall into a reasonable range. 
To demonstrate the effectiveness of our proposed MBCN, we conduct extensive offline and online experiments in a world-level video search engine.
The experimental results show that due to the powerful ability of MBCN to identify video quality issues, the ranking ability of the video retrieval system has been significantly improved.
We also conduct a series of detailed experimental analyses to verify that all four evaluation branches play a positive role.
Besides that, for emerging low-quality AI-generated videos, the recognition accuracy of MBCN also improves significantly compared to the baseline.

\end{abstract}





\maketitle

\section{Introduction}

Large industrial video retrieval systems like Google and Baidu must consider multiple characteristics in terms of texts and videos, such as relevance, freshness, and quality. Among them, video quality assessment (VQA) plays a vital role in accurately identifying low-quality videos. 
Improving the accuracy of VQA is very important, as it can encourage video retrieval systems to prioritize high-quality videos and enhance users' video search experience.

A substantial body of research in academia has already been dedicated to the problem of VQA~\cite{hosu2017konstanz, sinno2018large, ying2021patch, wang2019youtube}.
Many of these studies primarily assess low-level visual quality issues, including motion blur, flicker, abrupt scene transitions, and artifacts~\cite{Wu_Chen_Liao_Hou_Sun_Yan_Lin_2022, zhao2023zoom, Ying_Maniratnam_Ghadiyaram_Bovik_2020, Zhang_Wu_Sun_Tu_Lu_Min_Chen_Zhai_2023, Li_Jiang_Jiang_2019, Li_Jiang_Jiang_2021}.
Other studies focus on higher-level abstract quality issues, such as emotional value, aesthetics, and logical correctness~\cite{wu2022disentangling, wu2023exploring, he2024cover, wu2023towards}.
With the advancement of artificial intelligence (AI)~\cite{achiam2023gpt, ho2020denoising, touvron2023llama, videoworldsimulators2024}, more and more AI-generated videos emerge, and researchers gradually begin to pay attention to the authenticity of these AI-generated videos~\cite{unterthiner2019fvd, huang2024vbench, kou2024subjective, lu2024aigc}. 
However, AI-enhanced video editing tools~\cite{fried2019text, huh2023avscript, laput2013pixeltone, lin2023identifying, wang2024lave} are also rapidly developing, these tools automatically create stories, retrieve relevant shots, and splice video clips based on user instructions. 
With the advantages of simple operation and relatively low cost, these new tools can produce massive videos in a very short time and quickly flood into the Internet, posing new challenges to video quality assessment in the real world.

\begin{figure}[t]
  \centering
  \includegraphics[width=\linewidth]{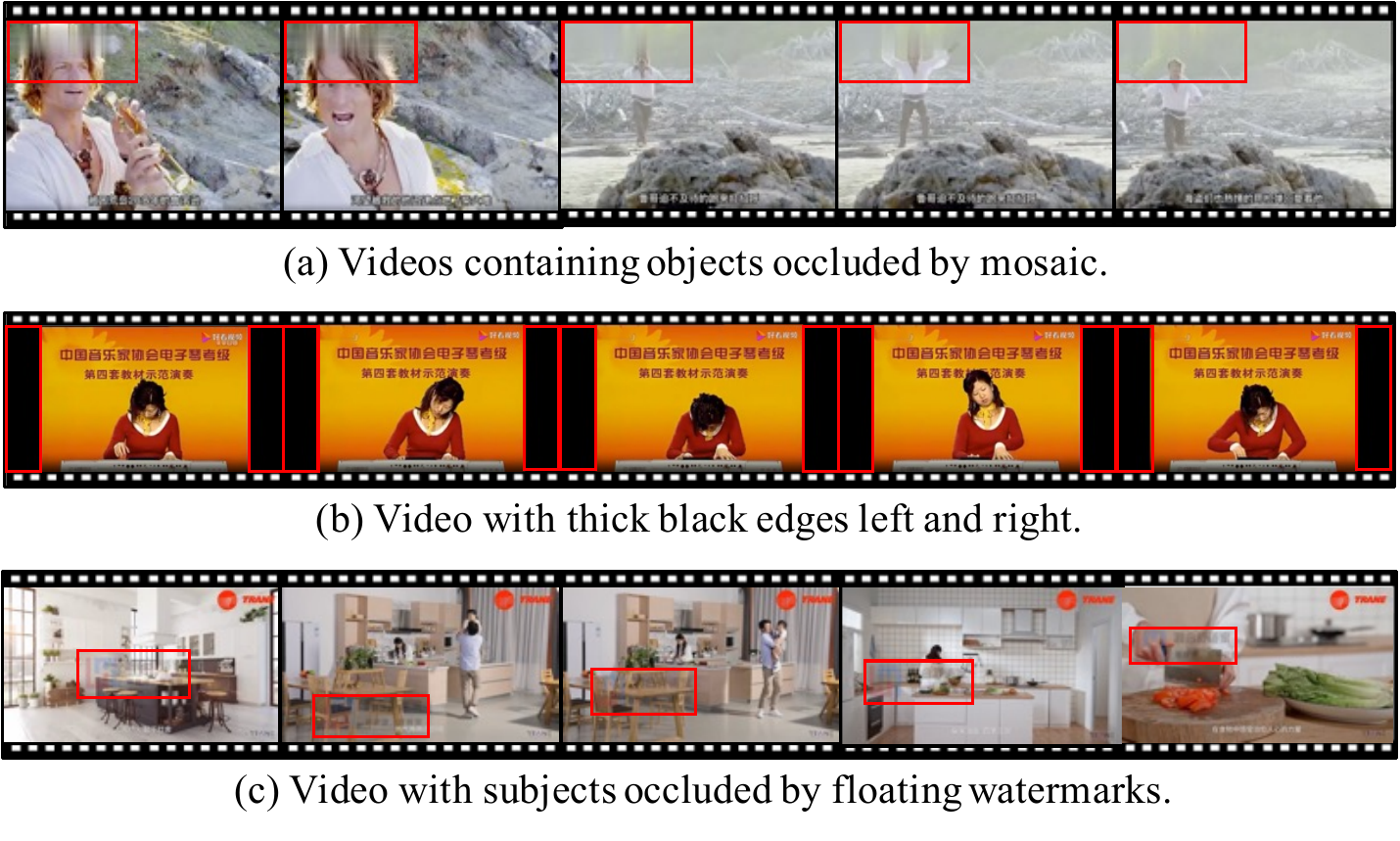}
  \caption{Traditional visual-related low-quality characteristics in industrial video retrieval systems.}
  \Description{A woman and a girl in white dresses sit in an open car.}
  \label{fig:genral_quality}
\end{figure}

Before achieving a sound VQA system, we first systematically introduce the typical characteristics of low-quality videos in industrial video retrieval systems.
Generally, traditional video quality issues are mainly divided into visual-related and text-related categories.
(1) \textbf{Visual-related low-leval}: It encompasses a series of low-level quality problems, such as mosaics, black boxes, and watermarks.
For example, the mosaic in Figure~\ref{fig:genral_quality}(a) partially obscures the face, the black box in Figure~\ref{fig:genral_quality}(b) occupies a portion of the frame, and the watermark in Figure~\ref{fig:genral_quality}(c) overlaps with objects in the video. It is obvious that all these problems affect the user's viewing experience.
(2) \textbf{Textual-related low-quality}: The main manifestations of such problems include weak correlation between video titles and Optical Character Recognition (OCR) content, and poor readability caused by grammatical errors in video titles. This type of problem is not conducive to video retrieval, and it is also difficult for users to rely on video titles to find desired videos.

\begin{figure}[t]
  \centering
  \includegraphics[width=\linewidth]{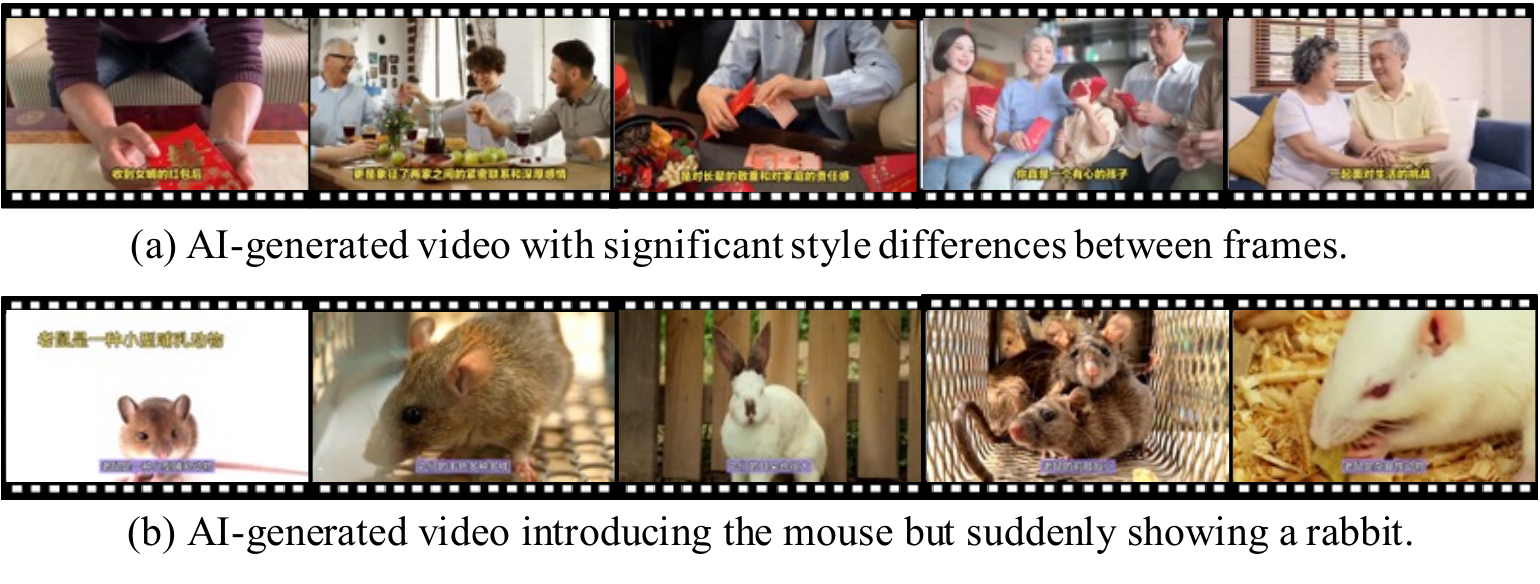}
  \caption{Characteristics of AI-generated low-quality videos in industrial video retrieval systems.}
  \label{fig:aigc_quality}
\end{figure}

With the development of TTV (text-to-video) technology, AI-generated videos have flooded into video retrieval systems, and a considerable number of videos are products of AI-enhanced video editing tools. They retrieve video clips from multiple videos based on natural language and then edit or splice these video clips together.
However, many videos generated by AI-enhanced tools are quite unsatisfactory. These AI-generated videos exhibit different low-quality features in both visual and textual dimensions compared to traditional user-generated videos.  
(1) \textbf{Frame incoherence}: 
Since many videos are spliced from multiple video clips, they tend to have poor frame coherence.
Figure~\ref{fig:aigc_quality}(a) shows several frames in an AI-generated video, and it can be seen that there are significant differences in the content and style of each frame, resulting in a strong sense of video splicing.
(2) \textbf{Frame-Text mismatch}: Another problem with these AI-generated videos is the poor relevance between frame images and video text. The title and OCR content of the video in Figure~\ref{fig:aigc_quality}(b) are both dedicated to introducing mice, but some frames show other animals, such as rabbits and locusts, which significantly affects users' trust in these videos.
Despite significant progress in the VQA field, these video quality issues faced in industrial environments have never been studied.

To effectively address practical challenges in the industrial system, we propose \textbf{M}ulti-\textbf{B}ranch \textbf{C}ollaborative learning \textbf{N}etwork (\textbf{MBCN}):
(1) \textit{Multimodal Encoder}: It first encodes the video frame sequence and its text. Considering the mutual relationship between video frames,
we utilize an additional transformer module to capture the temporal information between frames.
(2) \textit{Multi-Branch Quality Assessment}: 
This module contains four assessment branches corresponding to the four video quality problems we reviewed above. 
Specifically, in the Video-Text Matching Assessment Branch, we measure the relevance of video and its associated text with multi-grained image-text similarity.
The Frame Coherence Assessment Branch considers the coherence between video frames through inter-frame differences and differences between each frame and the whole video theme. 
The Frame Quality Assessment Branch takes frame-related information to model low-level visual quality issues such as mosaics, black boxes, and watermarks. 
Similarly, the Text Quality Assessment Branch utilizes all text semantic information to assess text quality issues such as poor readability.
(3) \textit{Squeeze-and-Excitation Aggregation}: 
It adopts the Squeeze-and-Excitation operation to adaptively model the influence of video content on different branches, dynamically addressing video quality issues in different scenarios within large-scale industrial systems.
(4) \textit{Optimization}: 
We integrate point-wise and pair-wise optimization objectives to ensure the predicted scores are stable and fall into a reasonable range. It makes the output score easily compatible with the ranking signals of other modules (such as relevance, freshness, and authority) in video retrieval systems.

\begin{figure*}[!ht]
    \centering
    \includegraphics[width=1\textwidth]{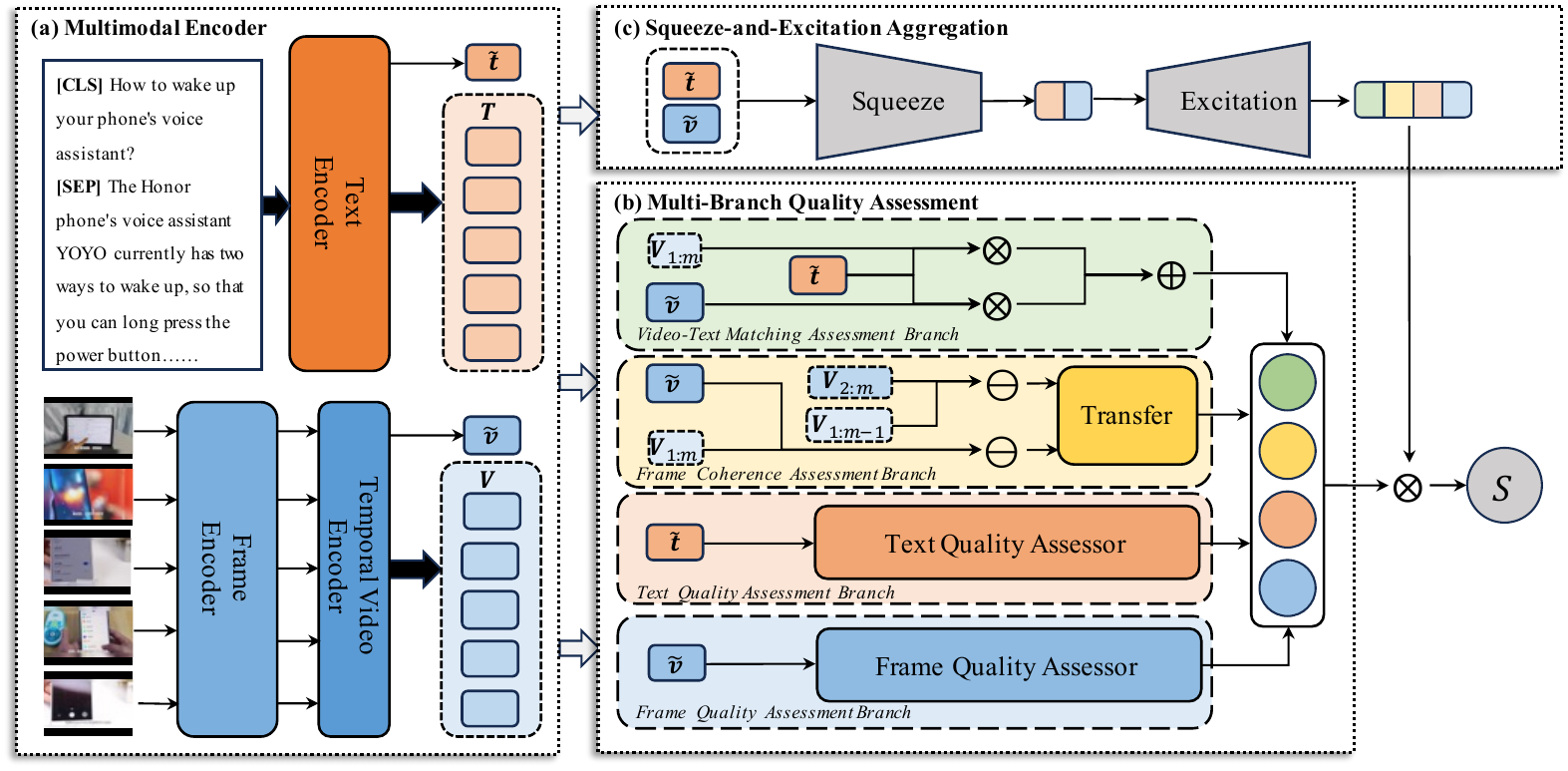}
    \caption{Illustration of the Multi-Branch Collaborative Learning Network (MBCN). It takes the text and frame images of the video as inputs to obtain text, frames, and video representations, where a frame encoder and a temporal encoder are combined as the frame encoder. Subsequently, four assessment branches are carefully designed to adapt to the four characteristics of low-quality videos in industrial video retrieval systems. Lastly, we perform a weighted aggregation of the various branches to dynamically address video quality issues in different scenarios with a squeeze-and-excitation mechanism.}
    \label{fig:framework}
    \vspace{-1em}
\end{figure*}

We conduct extensive offline and online experiments in a large-scale video search engine. 
Extensive experimental results demonstrate the effectiveness of our proposed method, and show a significant improvement in the ranking ability of the video retrieval system due to MBCN's stronger ability to identify video quality issues.
In addition, we also conduct a series of detailed experimental analyses. 
We validate that the four carefully designed evaluation branches all positively affect the final result.
Compared to the baseline, MBCN's predicted score is more discriminative and reasonable.
It is worth mentioning that for emerging low-quality AI-generated videos, the recognition accuracy of MBCN also has a significant improvement compared to baseline.

The main contributions are summarized as follows:

\begin{itemize}
    \item We systematically analyze the video quality issues in industrial video retrieval systems, formalizing them into four main problem types. To the best of our knowledge, it is the first time to systematically analyze the types of video quality issues in industrial video retrieval systems.
    \item Based on the analysis, we develop a Multi-Branch Collaborative Learning Network, MBCN, for video quality assessment in industrial video retrieval systems, which analyzes and assesses different aspects of video quality issues.
    \item We conduct extensive offline and online experiments to verify the effectiveness of the designed model. The results demonstrate that the proposed technique significantly enhances the performance of VQA. 
    \item We manually select a subset of AI-generated videos to evaluate the model’s performance in this specific scenario. The results indicate that the MBCN model notably improves the base model’s performance in assessing the quality of AI-generated videos.
\end{itemize}

\section{Methodology}

To build a large-scale video retrieval system, four types of low-quality videos should be given special attention: visual-related low-quality issues such as mosaic and black box, text-related low-quality problems caused by video titles and OCR content, as well as frame incoherence and frame-text mismatch in by emerging AI-generated videos.
In this section, we introduce MBCN, a Multi-Branch Collaborative Learning Network, to measure these low-quality issues individually. Figure~\ref{fig:framework} provides the whole workflow.

\subsection{Task Definition}
The task of video quality assessment is to measure the quality of a set of videos with their text \( \mathcal{N} = \{(v_i, t_i)\}_{i=1}^N \), where $t_i$ refers to the textual content of the video $v_i$. 
We aim to find a scoring function \( f(v, t) \to \mathbb{R} \) that maximizes the defined objective in equation (1),
\begin{equation}
    f^* = \arg \max_f \mathbb{E}_{\{v, t, y\}} [Y, F(v, t)],
\end{equation}
where \( f^* \) is the optimal scoring function, \( F(v, t) = \{f(v_i, t_i)\}_{i=1}^N \) is the set of video scores, and \( Y = \{y_i\}_{i=1}^N \) is the label set, with \( y_i \) representing the quality label. Typically, \( y_i \) is the graded quality label, which corresponds to the quality of \( (v_i, t_i) \) as \{\textit{bad}, \textit{fair}, \textit{good}, \textit{excellent}\} respectively.

\subsection{Multimodal Encoder}

\subsubsection{Text Encoder} 

Given a video $v_i$, we concatenate the title with the content extracted by the OCR tool as text $t_i$. 
Next, we utilize Chinese BERT\footnote{https://huggingface.co/google-bert/bert-base-chinese} as our text encoder to derive the textual representation in our video quality assessment system. 
Specifically, the text encoder is a transformer architecture that includes multi-head self-attention mechanisms and feed-forward neural networks. This transformer consists of 12 layers and 8 attention heads, with each query, key, and value vector having a dimensionality of 512. We apply a lower-cased byte pair encoding (BPE)~\cite{sennrich2015neural} method for tokenization. Before being fed into the text encoder, the token sequence is prefixed with [CLS] token. The output of the text encoder provides the global-level representation $\mathbf{\tilde{t}}_i \in \mathbb{R}^d$, and the token-level representations $\mathbf{T}_i = \{\mathbf{t}_{i,j}\}_{j=1}^{n_i} \in \mathbb{R}^{n_i \times d}$, where $n_i$ represents the length of the text $t_i$.

\subsubsection{Frame Encoder} 
\label{sec:Frame Representation}
For each video \( v_i \), we employ FFmpeg\footnote{https://ffmpeg.org} to extract \( m_i \) key frames, and all these frames are resized to 224x224, as the encoder requires.
In video quality assessment, in addition to keyframes, video covers also contain symbolic information, and their clarity is usually positively correlated with the quality of the entire video, so we resize and center-crop the video cover to obtain two cover frames and append them into the frame sequence.
To this end, an extended frame sequence \(V_i = \{v_{i,j}\}_{j=1}^{m_i+2}\) is formed. The frame encoder then processes these frames to obtain frame-level features, where the frame encoder is a standard Vision Transformer~(ViT), initialized with a public CLIP checkpoint~\cite{radford2021learning}. 
The architecture of ViT is similar to the encoder used in text representation, except that it introduces a visual tokenization process to convert frames into a visual token sequence. Finally, the visual token sequence, prepended with a [CLS] token, is then fed into frame encoder to achieve primary frame features.

\subsubsection{Temporal Video Encoder}

To consider the interaction between frames in a video, like tokens in a natural language sentence, we implement a temporal encoder to capture the interaction between different frames. Specifically, this encoder is a two-layer transformer block. A [CLS] token is added before all frame representations first, and then a temporal position embedding $\mathbf{P}$ is added to each frame to model the temporal relationship. Consequently, we obtain video representation $\{ \mathbf{\tilde{v}}_i \in \mathbb{R}^d, \mathbf{V}_i = \{ \mathbf{v}_{i,j} \}_{j=1}^{m_i} \in \mathbb{R}^{m_i \times d} \}$ through the temporal encoder, where $\mathbf{\tilde{v}}_i$ is the representation of the [CLS] token, representing the coarse-grained video feature, and $\mathbf{V}_i$ is fine-grained features for the frame sequence without video cover.

\subsection{Multi-Branch Quality Assessment}

Drawing from the analysis presented in the introduction, we establish four distinct branches to assess the representative types of quality issues.
Here, we propose a \textbf{M}ulti-\textbf{B}ranch \textbf{C}ollaborative learning \textbf{N}etwork (\textbf{MBCN}) that consist of four branches to evaluate the four perspectives of the problem.
First, we design a multi-grained image-text matching assessment branch to assess the video's alignment between image and text content. 
Second, we implement a frame coherence assessment branch to evaluate the coherence between video frames. 
Third, to evaluate various low-level quality issues in video frames, such as mosaics, black boxes, and watermarks, we create a frame quality assessment branch. 
Lastly, to examine the quality issues of video titles and OCR contents, we propose a text quality assessment branch.

\subsubsection{\textbf{V}ideo-\textbf{T}ext \textbf{M}atching \textbf{A}ssessment \textbf{B}ranch (\textbf{VTMAB})}

Due to the immaturity of AI-enhanced video editing tools or the limited availability of video libraries, some retrieved video clips may not strongly correlate with textual descriptions.
In this branch, we adopt multi-grained image-text similarity to evaluate the mismatch problem from both global and local levels.
Given the text representation $\mathbf{\tilde{t}}_i \in \mathbb{R}^d$, video representation $\mathbf{\tilde{v}}_i \in \mathbb{R}^d$, and frame representation sequence $\mathbf{V}_i \in \mathbb{R}^{m_i \times d}$, we first obtain a global video-text matching score $s_{vt}^g$ and local video-text matching score $s_{vt}^l$,
\begin{equation}
    s_{vt}^g = \mathbf{\tilde{v}}_i^{\top} \mathbf{\tilde{t}}_i,\ s_{vt}^l = \mathrm{mean}( \mathbf{V}_i^{\top} \mathbf{\tilde{t}}_i).
\end{equation}
Then, we take the mean of $s_{vtg}$ and $s_{vtl}$ as the video-text matching score $s_{vt}$, where $[\cdot:\cdot]$ means concatenation,
\begin{equation}
    s_{vt} = \mathrm{mean}([s_{vt}^g:\ s_{vt}^l]).
\end{equation}

\subsubsection{\textbf{F}rame \textbf{C}oherence \textbf{A}ssessment \textbf{B}ranch (\textbf{FCAB})}

The poor coherence between frames is also one of the main issues in AI-generated videos, including off-topic frame segments and excessive scene switching. Here, we introduce inter-frame differences to measure the coherence between frames.
Given the video-level feature $\mathbf{\tilde{v}}_i$ and the frame-level features $\mathbf{V}_i = \{ \mathbf{v}_{i,j} \}_{j=1}^{m_i}$, we remove the video covers from the frame sequence to focus on assessing video coherence. 
We obtain the global coherence matrix $\mathbf{M}_{fc}^g \in \mathbb{R}^{m_i \times d}$ and the local coherence matrix $\mathbf{M}_{fc}^l \in \mathbb{R}^{(m_i-1) \times d}$ by vector subtraction,
\begin{equation}
    \mathbf{M}_{fc}^g = \begin{bmatrix}
        \mathbf{\tilde{v}}_i - \mathbf{v}_{i,1} \\
        \vdots \\
        \mathbf{\tilde{v}}_i - \mathbf{v}_{i,m_i}
    \end{bmatrix},\quad
    \mathbf{M}_{fc}^l = \begin{bmatrix}
        \mathbf{v}_{i,2} - \mathbf{v}_{i,1} \\
        \vdots \\
        \mathbf{v}_{i,m_i} - \mathbf{v}_{i,m_i-1}
    \end{bmatrix}.
\end{equation}
We then use the transformation of the mean of $\mathbf{M}_{fc}^g$ and $\mathbf{M}_{fc}^l$ to obtain the score $s_{fc}$, where $\mathbf{W}_C \in \mathbb{R}^{d \times 1}$ is a learnable parameter,
\begin{equation}
    s_{fc} = \mathrm{Mean}(\begin{bmatrix}
        \mathbf{M}_{fc}^g \\
        \mathbf{M}_{fc}^l
    \end{bmatrix}^{\top}) \mathbf{W}_C.
\end{equation}

\subsubsection{\textbf{F}rame \textbf{Q}uality \textbf{A}ssessment \textbf{B}ranch (\textbf{FQAB})}

In addition to the semantic-level quality issues of AI-generated videos, we also focus on the traditional visual-related low-level quality issues. Recall that in Section~\ref{sec:Frame Representation}, we consider the center-cropped video cover for the clarity evaluation so that the video-level representation $\mathbf{\tilde{v}}_i$ is able to capture these low-level quality issues effectively. 
These complex and varied problems are challenging to model within a unified paradigm but are common enough to warrant attention.
In this branch, we employ an MLP behind the video representation to model these low-level quality issues and obtain the video quality score $s_{fq}$,
\begin{equation}
    s_{fq} = \mathrm{MLP}_F(\mathbf{\tilde{v}}_i).
\end{equation}

\subsubsection{\textbf{T}ext \textbf{Q}uality \textbf{A}ssessment \textbf{B}ranch (\textbf{TQAB})}

For text quality problems, we evaluate issues such as clickbait and content holes. Similar to the visual-related low-level quality problem, these textual-related quality issues are also trivial and difficult to capture. Here, we also apply an MLP to model and evaluate them. For a given text representation $\mathbf{\tilde{t}}_i$, we obtain the text quality score $s_{tq}$ as,
\begin{equation}
    s_{tq} = \mathrm{MLP}_T(\mathbf{\tilde{t}}_i).
\end{equation}

\subsection{Squeeze-and-Excitation Aggregation}

Through the Multi-Branch Quality Assessment Module, we evaluate various dimensions of video and text. However, different types of videos exhibit distinct quality issues, and it is necessary to adaptively integrate the scores of each branch together. For instance, low-quality AI-generated videos may suffer from poor frame coherence, while clickbait videos may encounter more severe image-text matching problems.
SENet~\cite{hu2018squeeze} is a neural network architecture that improves the model's representation ability and performance by introducing a Squeeze-and-Excitation module to readjust the feature representation of each channel adaptively.
Here, we draw the idea to design a Squeeze-and-Excitation Aggregation Module suitable for video assessment, which adaptively evaluates the proportion of different quality issues in the overall quality assessment according to the video content.
The formulation is as follows,
\begin{equation}
    \begin{aligned}
        \mathbf{z} &= \mathrm{softmax}( \delta( \begin{bmatrix} \mathbf{\tilde{v}}_i \\ \mathbf{\tilde{t}}_i \end{bmatrix} \mathbf{W}_S)^{\top} \mathbf{W}_E),\\
        s &= \sigma([s_{vt};\ s_{fc};\ s_{fq};\ s_{tq}] \mathbf{z}^{\top}),
    \end{aligned}
\end{equation}
where $\mathbf{W}_S \in \mathbb{R}^{d \times 1}$ and $\mathbf{W}_E \in \mathbb{R}^{2 \times 4}$ are learnable parameters used to squeeze and then extend the score vector. The function $\delta$ is the ReLU activation, and $\sigma$ is the Sigmoid function used to normalize the final quality score $s$.

\subsection{Optimization}

In the task definition, we classify video quality into four levels: \{\textit{bad}, \textit{fair}, \textit{good}, \textit{excellent}\}, with their corresponding soft labels as $\{0, 0.3, 0.6, 1\}$. Given the model output $f(v_i, t_i)$ and its soft label $\hat{y}_i$, a straightforward approach is to optimize the model using pointwise loss directly,
\begin{equation}
\mathcal{L}_{point} = \frac{1}{N} \sum_{i=1}^{N} (f(v_i, t_i) - \hat{y}i)^2,
\end{equation}
where $N$ is the batch size. 
However, solely minimizing the difference between the predicted scores and the soft labels may not make the model's scores as discriminative as we expect.
This approach fails to capture the correlations between different samples. 
Therefore, we introduce a pairwise loss to encourage the model to distinguish between different samples and ensure a more concentrated score distribution for similar samples,
\begin{equation}
\mathcal{L}_{pair} = \frac{1}{N} \sum_{y_i < y_j} \max(0, f(v_j, t_j) - f(v_i, t_i) + \tau),
\end{equation}
where $\tau$ is a margin hyper-parameter. The final loss is defined as a combination of the pointwise and pairwise losses,
\begin{equation}
\mathcal{L} = \alpha \mathcal{L}_{point} + (1 - \alpha) \mathcal{L}_{pair},
\end{equation}
where $\alpha$ is a weighting factor. By incorporating the pairwise loss, the model's quality score predictions become more stable and distinguishable. This combined loss function enables the model to learn both the absolute and relative quality of videos, improving its overall performance and robustness.

\section{Experiments}
To ensure the effectiveness of the proposed approaches, we conduct extensive offline and online experiments on a large-scale real-world retrieval system. 
This section details the experimental setup and presents several conclusions demonstrating that the proposed approaches are crucial to the video retrieval system.

\subsection{Dataset}
We evaluate our model with the dataset collected from the actual production environment. Specifically, we collect videos from the video retrieval system to ensure the authenticity and diversity of the dataset.
Next, the dataset is labelled on Baidu’s crowd-sourcing platform, where a group of experts are required to assign a quality label that varies from \{\textit{bad}, \textit{fair}, \textit{good}, \textit{excellent}\} for each video. 
Ultimately, our manually annotated dataset includes 110,502 videos for training and 15,068 videos for testing.

Here we briefly introduce the data annotation guidelines. 
Based on user's viewing experience, we divide video quality into four labels \{\textit{bad}, \textit{fair}, \textit{good}, \textit{excellent}\}. 
The \textit{bad} label indicates that the video has very serious quality issues and is unbearable; The \textit{fair} label indicates that the video has some minor flaws that slightly affect the viewing experience; The \textit{good} label means the video has no noticeable flaws and the overall viewing experience is satisfactory; The \textit{excellent} label indicates that the video is very well-produced and the overall viewing experience is highly satisfactory.

\subsection{Evaluation Metrics}
We employ the following evaluation metrics to assess the performance of the proposed model and the ranking system.

The \textbf{Positive-Negative Ratio} (PNR) is a commonly used evaluation metric for ranking tasks. In industrial retrieval scenarios, we use the PNR metric to measure the performance of ranking models, which helps us select better ranking models. For a list of N-ranked documents, the PNR is defined as the ratio of the number of pairs in correct order to the number of pairs in reverse order: 
\begin{equation}
PNR = \frac{\sum_{i,j \in [1,N]} \mathds{1}\{y_i > y_j\} \cdot \mathds{1}\{\tilde{y}_i > \tilde{y}_j\}}{\sum_{i,j \in [1,N]} \mathds{1}\{y_i > y_j\} \cdot \mathds{1}\{\tilde{y}_i < \tilde{y}_j\}},
\end{equation}
where the value of the indicator function $\mathds{1}\{x > y\}$ is 1 when $x > y$ and 0 otherwise. 

The \textbf{Area Under Curve} (AUC) is an important metric used to measure the ranking performance of a model. It represents the probability that the positive and negative samples predicted by the model are sorted correctly. Specifically, AUC is the area under the ROC curve, which is obtained by plotting the True Positive Rate and False Positive Rate at different classification thresholds. The model sorts all samples according to their predicted probabilities in this process. Thus, AUC is a measure of the model's ability to rank the samples. The formula is as follows,
\begin{equation}
    AUC = \int_{0}^{1} TPR(t) \: dFPR(t),
\end{equation}
\begin{equation}
    TPR(t) = \frac{TP(t)}{TP(t) + FN(t)},
\end{equation}
\begin{equation}
    FPR(t) = \frac{FP(t)}{FP(t) + TN(t)},
\end{equation}
where $TP$, $FP$, $FN$, and $TN$ denote the number of true positives, false positives, false negatives, and true negatives, respectively, at a threshold value of $t$.

The \textbf{Discounted Cumulative Gain} (DCG) is a metric used to measure the ability of information ranking, commonly employed in scenarios such as search engines and recommendation systems for assessing model performance. It takes into account the relevance of all retrieved documents and their positions in the retrieval results, and gives a higher weight to the relevant documents that are ranked first. Therefore, the ranking performance of the model can be evaluated more comprehensively. The formula is as follows:
\begin{equation}
    DCG_N = \sum_{i=1}^{N} \frac{G_i}{\log_2(i + 1)},
\end{equation}
where $G_i$ represents the matching degree of the document at position $i$. The more relevant the document is to the query, the larger $G_i$ is. $\frac{1}{\log_2(i + 1)}$ is the value of position i. The later the document is sorted, the lower the value it can produce. $N$ represents the $DCG$ value of topN, e.g. $DCG_2$ represents the dcg value generated by the top2 documents in the sort, $DCG_4$ represents the dcg value generated by the top4 documents in the sort.

The \textbf{Good vs. Same vs. Bad} (GSB) is a metric judged by professionally trained annotators. For a user query, we provide the annotators with both System A and System B search results. The annotators need to independently give Good (result A is better than B), Bad (result B is better than A), and Same (they are equally good or bad), by judging the quality of the search results, and their relevance to the query. To quantify human evaluation, we use a unified metric to describe Good, Same, and Bad denoted $\Delta$GSB,
\begin{equation}
    \Delta GSB = \frac{\# \text{Good} - \# \text{Bad}}{\# \text{Good} + \# \text{Same} + \# \text{Bad}}.
\end{equation}

\subsection{Competitor System}
To demonstrate the effectiveness of our method, we compare it with multiple strong baselines.
In addition, we also compare different variants of the proposed method.
Given the high costs associated with deploying suboptimal models, we limit our online experiments to comparing the proposed model with the state-of-the-art CLIP-based model, which serves as the baseline currently deployed in the video retrieval system.
Based on past experience, the conclusions of online experiments are consistent with the offline results.

\begin{itemize}
\item Base: The base model is a 12-layer CLIP-based video quality assessment model, fine-tuned with
human-labeled videos. It directly feeds text and video representations into a feedforward neural network for prediction.
\item COVER~\cite{he2024cover} and DOVER~\cite{wu2023exploring}: They are both strong baselines in the VQA field, primarily focusing on low-level quality issues such as motion blur and flicker, as well as higher-level abstract issues including aesthetics and logical correctness.
\item X-CLIP~\cite{ni2022expanding}: X-CLIP is implemented based on CLIP and serves as a strong baseline in the field of video recognition. We additionally test its performance on our dataset.
\item MBCN: The main model of this paper. Multi-Branch Collaborative learning Network is used to better handle complex scenarios in video quality assessment.
\item Model variants: A series of model variants created by removing branches from each of the four dimensions of MBCN.
\end{itemize}

\subsection{Implementation Details}
For the text encoder, we use the bert-base-chinese version. Specifically, the number of encoder layers $L = 12$, the size of the hidden size $H = 768$, the number of heads of multi-head attention $A = 12$, and the total number of parameters is 110M. 
For the frame encoder, we use the VIT visual encoding model, where the patch size is 32, the width is 768, the number of layers is 12, and the number of heads for multi-head attention is 12. 
For the video encoder, we use the transformer model for multi-frame aggregation based on frame encoding, the number of encoder layers is 2, and the hidden size is 512. 
For each video, we use ffmpeg to extract key frames, up to a maximum of 20 frames, and resize each frame to 224*224.
For the generic hyperparameters, we set the learning rate to 0.0001, dropout to 0.1, batch size to 128, $\alpha$ to 0.5, and used Adam as the optimizer. 
MBCN and the online baseline employ the same text encoder and video encoder, and they obtain text and video representations in the same way. Their parameter scales are comparable.
Finally, we train by freezing the text encoder and frame encoder parameters. Thus, training is relatively low-cost, requiring only a 40G NVIDIA A100.

\subsection{Offline Experimental Results}

\begin{table}[t]
    \caption{Offline comparison of different models. 
``*'' represents the results obtained by us after fine-tuning it on our training set.}
    \label{tab:offline}
    \begin{tabular}{l|cc}
    \toprule
    Model & PNR & AUC \\
    \midrule
    Base & 5.380 & 0.829 \\
    COVER*~\cite{he2024cover} & 2.931 & 0.725 \\
    DOVER*~\cite{wu2023exploring} & 2.804 & 0.712 \\
    X-CLIP*~\cite{ni2022expanding} & 5.139 & 0.830 \\
    \midrule
    MBCN & \textbf{5.710} & \textbf{0.855} \\
    w/o VTMAB & 5.588 & 0.849 \\
    w/o FCAB & 5.500 & 0.845 \\
    w/o FQAB & 5.530 & 0.847 \\
    w/o TQAB & 5.633 & 0.850 \\
    \bottomrule
    \end{tabular}
\end{table}

In the offline evaluation, we report the PNR and AUC results on the test set. Note that the AUC result here represents \{\textit{bad}, \textit{fair}\} vs \{\textit{good}, \textit{excellent}\}, which represents the model’s ability to distinguish between positive and negative samples. Table 1 shows the results, where some key findings are observed:

\begin{itemize}
\item Among all the models, our MBCN achieves the best performance. 
COVER~\cite{he2024cover} and DOVER~\cite{wu2023exploring}, strong VQA baselines, are outperformed by MBCN due to their limitations in model input, modeling objectives, and structural design, hindering their efficacy in real industrial scenarios. 
Additionally, although X-CLIP~\cite{ni2022expanding}, a strong baseline in the video recognition field, considers both text and frame images as input, it does not explicitly model the four types of video quality issues present in real industrial scenarios, resulting in a noticeable gap compared to MBCN.
\item After separately removing the four assessment branches of MBCN, the results have declined to varying degrees, which shows that our four assessment branches can handle different types of quality problems and play a significant role in the final improvement. By integrating the four assessment branches together, our model achieved the best results. Compared to the baseline, it improved the PNR and AUC results by 6.12\% and 3.13\%, respectively.
\item From the AUC and PNR metrics, it can be observed that our model not only enhances the ability to distinguish positive and negative samples, but also improves the overall ranking ability. In other words, compared to the baseline, our model can not only identify the positive samples, but also better distinguish between good and excellent videos, thereby presenting videos of higher quality to users.
\item According to the ablation experiments on the four assessment branches, it can be seen that the removal of FCAB has the most significant decrease in effectiveness, followed by FQAB, and finally, VTMAB and TQAB. The results indicate that the main problem in industrial video retrieval systems is still related to the visual image, such as poor image coherence and low-quality image issues (black boxes, mosaics, watermarks, etc.). However, this does not mean that image text matching and text issues are not necessary. When we remove the VTMAB and TQAB separately, the decline in performance is also quite noticeable.
\end{itemize}

\subsection{Online Experimental Results}

\begin{table}[t]
\caption{Performance improvements of online testing. ``*'' indicates the statistically significant improvement (t-test with $p < 0.05$ over the baseline).}
\label{tab:ablation}
\begin{tabular}{l|cc|cc}
\toprule
\multirow{2}{*}{Model} & \multicolumn{2}{c|}{$\bigtriangleup GSB$} & \multicolumn{2}{c}{$\bigtriangleup DCG$} \\ 
\cmidrule(lr){2-3}\cmidrule(lr){4-5}
 & Random & Long-Tail & $\bigtriangleup DCG_2$ & $\bigtriangleup DCG_4$ \\ \midrule
Base & -      & -         & -    & -    \\
MBCN & \textbf{+7.00}\%* & \textbf{+8.50}\%*  & \textbf{+3.55}\%* & \textbf{+4.18}\%* \\ 
\midrule
w/o VTMAB & +5.50\%* & +4.50\%  & +2.26\% & +3.45\%* \\ 
w/o FCAB & +4.00\% & +3.00\%  & +1.43\% & +1.35\% \\
w/o FQAB & +4.50\% & +3.50\%  & +2.03\% & +1.72\% \\
w/o TQAB & +6.00\%* & +6.00\%*  & +3.24\%* & +3.09\%* \\
\bottomrule
\end{tabular}
\begin{flushleft}
\centering
\end{flushleft}
\end{table}

To investigate the effectiveness of our new methods in the real-world video search engine, we deployed the proposed model on Baidu’s search engine, a world-level search system, and compare it with the baseline model in the real production environment.
Table 2 reports the performance comparison between different models regarding $\bigtriangleup GSB$ and $\bigtriangleup DCG$. 
The conclusions can be drawn:

\begin{itemize}
\item The full strategy model MBCN has the most significant improvement in all metrics, which proves that the integration of the four multidimensional assessment branches can be more helpful. Meanwhile, by removing each of the four assessment branches, there are different degrees of decrease in all online metrics, which proves that each assessment branch is extremely useful and can correspond to the offline conclusions.
\item In addition, we find that our model has a more noticeable effect advantage under long-tail queries (i.e., the search frequency of the query is lower than 10 per week). Specifically, the complete model MBCN achieves a relative improvement of 8.5\% in GSB under long-tail queries. At the same time, removing each assessment branch also decreases significantly under long-tail queries. This is because there are more low-quality and fewer high-quality videos under long-tail queries. Our proposed model can identify low-quality videos more accurately, thereby benefiting the high-quality videos to be ranked up, which is very helpful in improving the user's video search experience.
\item Compared to the base system, the proposed approach improves 3.55\% and 4.18\% in $DCG_{2}$ and $DCG_{4}$ metrics, respectively, prompting a significant improvement in the online system's ranking ability. 
It indicates that the new retrieval system is more inclined to rank out high-quality videos compared to the baseline due to the more accurate recognition of video quality by our model.

\end{itemize}

\subsection{Detailed Experimental Analysis}
In order to analyze the source of the effectiveness of our methods and better understand the characteristics of the proposed methods, we conduct a series of detailed experiments.

\subsubsection{Effects of Four Assessment Branches}

\begin{table}[t]
    \caption{The mean predicted logits of each assessment branch under different labels.}
    \label{tab:logits}
    \centering
    \begin{tabular}{l|cccc}
    \toprule
    logits & bad & fair & good & excellent \\ 
    \midrule
    VTMAB  & -0.91 & -0.87 & -0.76 & -0.69 \\ 
    FCAB & -3.19 & -2.41 & 1.20  & 5.82  \\ 
    FQAB      & -0.23 & -0.20 & 0.01  & 0.31  \\ 
    TQAB       & 0.21  & 0.32  & 0.65  & 0.67  \\ 
    \bottomrule
    \end{tabular}
\end{table}

To validate the effectiveness of our model framework further, we analyze the four multidimensional assessment branches.

Firstly, to demonstrate the discriminative power of the predicted values of each branch, we statistics the mean values of their logits under different labels in the test set.
As shown in Table 3, the predicted logits of each branch become more significant as the label increases, proving that their predicted values are positively correlated with the labels, which indicates that each branch can solve different types of problems in video quality modelling scenarios.
Secondly, in addition to calculating the average of the predicted values, we also conduct a quantitative analysis of the roles of each assessment branch to more intuitively demonstrate their respective functions. Table 4 shows the PNR of each branch's predicted logits with respect to the labels, and it can also be seen that all branches play a positive role. The FCAB has the highest discriminative ability, followed by the FQAB, and finally, the VTMAB and TQAB. This conclusion is related to the distribution of quality problem types in the current video retrieval system, and is basically consistent with the conclusion in Table 1.

\begin{table}[t]
    \caption{PNR of predicted logits from different assessment branches on the test set.}
    \label{tab:PNR}
    \centering
    \begin{tabular}{l|cccc}
    \toprule
     & VTMAB & FCAB & FQAB & TQAB \\ 
    \midrule
    PNR & 2.569 & 5.58 & 3.868 & 1.515 \\
    \bottomrule
    \end{tabular}
\end{table}

\subsubsection{Analysis of Model Scoring Distribution}

\begin{figure}[t]
  \centering
  \includegraphics[width=0.8\linewidth]{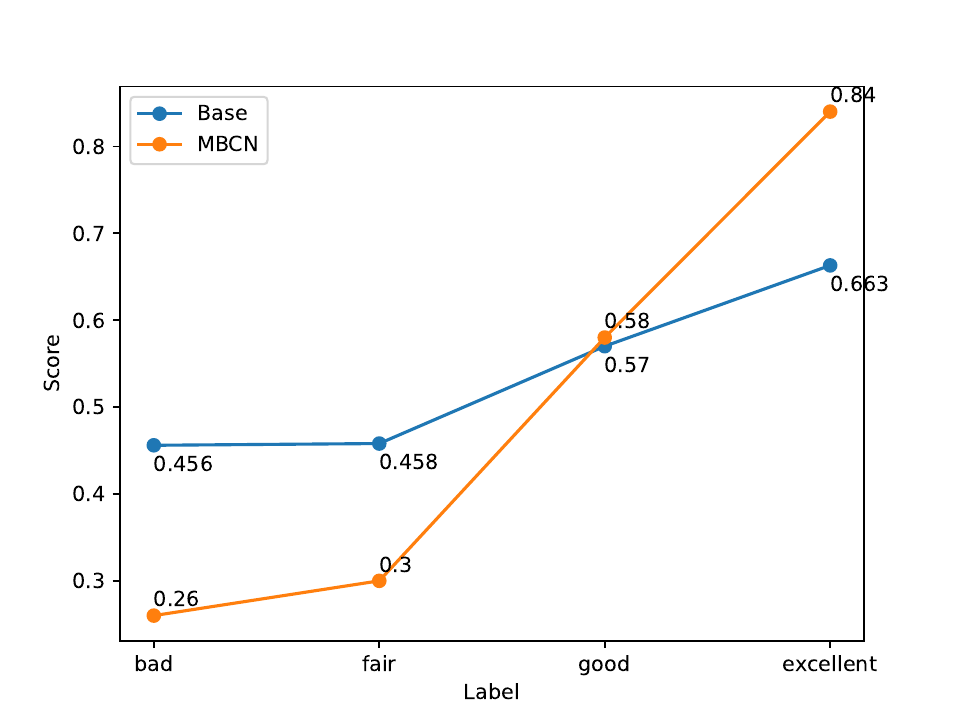}
  \caption{Comparison of average model prediction scores under different labels.}
  \Description{A woman and a girl in white dresses sit in an open car.}
  \label{fig:genral quality}
\end{figure}

Figure 4 shows the mean prediction scores of the baseline model and our model under each label on the test set. Compared to the baseline, our model's predicted scores are more discriminative and reasonable.
For example, the difference between the mean scores of the baseline for good and excellent video is only 0.09, while that of our model is 0.26. This difference in the distribution of the scores enables our model to distinguish high-quality videos from average-quality videos more clearly, which effectively improves the ranking effect of online video retrieval systems. We believe this is mainly due to the following two aspects:
\begin{itemize}
\item From the perspective of model design, our multidimensional assessment branches can handle a more comprehensive range of quality problem types, and each assessment branch is designed to solve specific scenario problems. As discussed in section 3.7.1, each branch plays a positive role in the final effect.
\item From the perspective of loss function optimization, compared to using only pair-wise loss for the baseline, our model's use of a mixture of point-wise and pair-wise loss can make the predicted scores more stable and encourage them to fall into a more meaningful range. This also makes our model score easily compatible with the ranking signals of other modules (such as relevance, freshness, and authority) in real-world video ranking systems.
\end{itemize}

\subsubsection{Evaluation on the AI-generated Video Test Set}

As mentioned in the previous sections, it is challenging to identify low-quality AI-generated videos in industrial video retrieval systems accurately.
To further demonstrate the effectiveness of MBCN, we manually select 200 low-quality AI-generated videos with labels of bad or fair and analyze the scores of MBCN and baseline.

Our model’s prediction layer uses the sigmoid activation function, so we set 0.5 as the threshold. Furthermore, we also evaluate both models’ accuracy metrics with a threshold of 0.4.
As shown in Table 5, in terms of mean score, the mean prediction score of MBCN on this low-quality AI-generated video test set is 0.175 lower than the baseline model, indicating a lower and more reasonable prediction score.
At the same time, we identify videos with model scores below the threshold $t$ (0.4 and 0.5) as low-quality AI-generated videos, and compare the performance of MBCN and baseline model by comparing their recognition accuracy.
From the results, it can be seen that as the threshold is set lower, the accuracy of MBCN is more significantly improved than the baseline. E.g., when the threshold is set to 0.4, the accuracy of MBCN is more than 20\% higher than the baseline. 
This indicates that MBCN has a lower prediction score for low-quality AI-generated videos than the baseline and can more accurately identify such videos.

\begin{table}[t]
    \caption{Comparison of ACC and mean scores of models on AI-generated video test set, where $t$ refer the threshold for determining low-quality videos.}
    \label{tab:aigc_test}
    \centering
    \begin{tabular}{l|ccc}
    \toprule
     & ACC ($t$=0.5) & ACC ($t$=0.4) & mean\_score \\ 
    \midrule
    Base & 84.0\% & 64.5\% & 0.401 \\ 
    MBCN & 93.5\% & 88.5\% & 0.226 \\ 
    \bottomrule
    \end{tabular}
\end{table}

\section{Related Works}

\subsection{Pretrained Vision-Language Models}

Pretrained vision-language models have emerged as a significant research direction in computer vision and natural language processing, achieving remarkable progress in recent years.
By jointly pre-training on large-scale image and text datasets, these models demonstrate outstanding performance across various downstream tasks, including image caption~\cite{mokady2021clipcap, stefanini2022show, hessel2021clipscore, vinyals2015show}, visual question answering~\cite{pan2023retrieving, song2022clip, antol2015vqa}, and cross-modal retrieval~\cite{xia2023clip, zhen2019deep}.
These models aim to improve multimodal data comprehension by capturing semantic associations between images and text. Common pretraining methods include contrastive learning and masked language modelling.
Contrastive learning methods, such as CLIP~\cite{radford2021learning} and ALIGN~\cite{jia2021scaling}, enable models to effectively understand cross-modal information by maximizing the similarity between correct image-text pairs while minimizing the similarity between incorrect pairs. 
Masked language model-based methods, such as VisualBERT~\cite{li2019visualbert} and VilBERT~\cite{lu2019vilbert}, enhance text understanding by incorporating visual information or generating richer semantic representations.
Some emerging methods, such as ALBEF~\cite{li2021align} and BLIP~\cite{li2022blip}, introduce innovations in pretraining strategies. ALBEF enhances multimodal representation learning by aligning visual and linguistic features and fusing them. Conversely, BLIP shows great potential by guiding the learning process and progressively enhancing the model’s performance on vision and language tasks.
As a pre-trained model with a simple structure yet powerful performance, CLIP has been widely utilized as a foundational model for numerous multimodal tasks~\cite{xia2023clip, song2022clip, mokady2021clipcap}.
However, only a few studies utilize CLIP’s cross-modal understanding ability to assist VQA~\cite{xing2024clipvqa}. In this paper, we explore using CLIP as a bridge to connect text and vision for VQA.

\subsection{Video Quality Assessment}

The primary goal of VQA is to accurately predict the perceived quality from the perspective of human viewers. It plays a vital role in meeting the promised quality of service (QoS) and in improving the end user’s quality of experience (QoE)~\cite{chikkerur2011objective}.
Numerous studies have leveraged hand-crafted features to capture quality perception features in videos~\cite{Saad_Bovik_Charrier_2014, Mittal_Saad_Bovik_2016, Korhonen_2019, Tu_Wang_Birkbeck_Adsumilli_Bovik_2021}. 
However, with the widespread adoption of deep learning and the emergence of large-scale VQA datasets~\cite{hosu2017konstanz, Wang_Inguva_Adsumilli_2019, Ying_Maniratnam_Ghadiyaram_Bovik_2020, Zhang_Wu_Sun_Tu_Lu_Min_Chen_Zhai_2023}, researchers have been increasingly drawn to its powerful capabilities and flexibility, resulting in a gradual shift in research focus towards deep learning-based VQA.
Some of these studies begin with low-level technical quality issues in videos, such as motion blur, flicker, and artefacts, and design a series of models to capture quality-aware features~\cite{Wu_Chen_Liao_Hou_Sun_Yan_Lin_2022, zhao2023zoom, Ying_Maniratnam_Ghadiyaram_Bovik_2020, Zhang_Wu_Sun_Tu_Lu_Min_Chen_Zhai_2023, Li_Jiang_Jiang_2019, Li_Jiang_Jiang_2021}.
DisCoVQA~\cite{Wu_Chen_Liao_Hou_Sun_Yan_Lin_2022} leverages the powerful temporal modelling capabilities of transformer architecture to differentiate temporal variations better and capture temporal distortions.
Zoom-VQA~\cite{zhao2023zoom} proposes a hierarchical structure to capture video quality issues at the patch, frame, and clip levels. 
Some other studies focus on higher-level abstract quality issues, such as emotional value, aesthetic and logicality, by guiding different types of encoders to capture abstract features in videos~\cite{wu2022disentangling, wu2023exploring, he2024cover, wu2023towards}. 
Dover~\cite{wu2022disentangling} employs disentangled representation learning to distinguish technical and aesthetic features in video representations. It then designs two distinct evaluators to assess aesthetic and technical quality.
However, the industry faces more complex quality problems, such as watermarking, video coherence, and image-text consistency, which cannot be easily classified under the paradigms studied in academia.

\subsection{AI-Enhanced Video Editing}

The rapid development of generative artificial intelligence technology~\cite{achiam2023gpt, ho2020denoising, touvron2023llama}, particularly in image~\cite{ramesh2021zero} and video~\cite{videoworldsimulators2024} generation, enables models to generate specific content guided by natural language.
In the field of video creation, numerous recent studies have incorporated natural language into the instructions of video editing tools. This allows AI-driven tools to automatically create and edit parts of the video content based on natural language~\cite{fried2019text, huber2019b, huh2023avscript, pavel2020rescribe, laput2013pixeltone, chang2019design, lin2023identifying, wang2024lave}.
LAVE~\cite{wang2024lave} designs a video editing agent based on Large Language Models (LLMs), incorporating a series of AI-enhanced functions for video editing. This agent can automatically create stories, retrieve relevant shots, and edit timelines according to user commands.
The proliferation of similar editing tools has resulted in a significant increase in videos co-created by humans and AI, which are now prevalent in video retrieval systems.
Although some researchers have recently specialized in assessing the quality of AI-generated videos~\cite{carreira2017quo, unterthiner2019fvd, huang2024vbench, kou2024subjective, lu2024aigc}, their primary focus has been on end-to-end text-to-video generation models, rather than the numerous videos co-created by humans and AI found in industrial video retrieval systems.

\section{Conclusion}

In this study, we conduct a systematic analysis of low-quality videos in current industrial video retrieval systems and then successfully deploy a powerful video quality assessment module centred around the innovative Multi-Branch Collaborative Leaning Network (MBCN).
Based on the different characteristics of low-quality videos, we design different branches to model relevant specialities from different dimensions. Among them, the video-text matching and frame coherence assessment branches are specifically proposed to focus on emerging issues of AI-generated videos.
The evaluation results of each branch are dynamically aggregated through an aggregation module, ensuring a comprehensive assessment of video quality. 
Extensive experimental results confirm the significant advantages of MBCN in industrial video retrieval systems, further enhancing performance on AI-generated videos and improving the stability of model scoring. These findings highlight the practical value of MBCN in real-world applications.


\bibliographystyle{ACM-Reference-Format}
\bibliography{ref}

\appendix

\end{document}